\documentclass{article}

\usepackage{booktabs} 
\usepackage{lipsum}
\usepackage{multicol}

\usepackage{epsfig,natbib,amsmath}
\usepackage[ruled]{algorithm2e} 

\SetAlFnt{\small}
\SetAlCapFnt{\small}
\SetAlCapNameFnt{\small}
\SetAlCapHSkip{0pt}
\IncMargin{-\parindent}

\begin{document}
\title{Theoretical Impediments to Machine Learning \\
With Seven Sparks from the Causal Revolution} 

\author{Judea Pearl \\ 
University of California, Los Angeles \\
Computer Science Department\\
Los Angeles, CA, 90095-1596, USA\\
judea@cs.ucla.edu }


%
%

%
%


\maketitle
\begin{abstract}
Current machine learning systems operate, almost exclusively, 
in a statistical, or model-free mode, which entails severe 
theoretical limits on their power and performance. 
Such systems cannot reason about
interventions and retrospection and, therefore, cannot serve as
the basis for strong AI. To achieve human level
intelligence, learning machines need the guidance of
a model of reality, similar to the ones used in
causal inference tasks.  To demonstrate the essential role of 
such models, I will present a summary of seven tasks which
are beyond reach of current machine learning systems and which
have been accomplished using the tools of causal modeling.
\end{abstract}

\subsection*{Scientific Background} 

If we examine the information that drives machine 
learning today, we find that it is almost entirely 
statistical. In other words, learning machines improve 
their performance by optimizing parameters over a stream of 
sensory inputs received from the environment. 
It is a slow process, analogous in many respects 
to the natural selection process that drives Darwinian evolution. 
It explains how species like eagles and snakes have developed
superb vision systems over millions of years. 
It cannot explain however the super-evolutionary
process that enabled humans to build eyeglasses and telescopes 
over barely one thousand years. What humans possessed that 
other species lacked was a mental representation, a
blue-print of their environment which they could manipulate 
at will to {\em imagine} alternative hypothetical environments 
for planning and learning. Anthropologists like  N.\ Harari, 
and S.\ Mithen are in general agreement that the decisive 
ingredient that gave our Homo sapiens ancestors the 
ability to achieve global dominion, about 40,000 
years ago, was their ability to choreograph a mental representation
of their environment, interrogate that representation, distort
it by mental acts of imagination and finally
answer ``What if?'' kind of questions. 
Examples are interventional questions: ``What if I act?'' 
and retrospective or explanatory questions: 
``What if I had acted differently?'' 
No learning machine in operation today can answer such questions 
about interventions not encountered before, say, ``What if we ban cigarettes.''
Moreover, most learning machines today do not provide a representation 
from which the answers to such questions can be derived.

I postulate that the major impediment to achieving accelerated 
learning speeds as well as human level performance should be 
overcome by removing these barriers and equipping 
learning machines with causal reasoning tools. This postulate 
would have been speculative 
twenty years ago, prior to the mathematization of counterfactuals. 
Not so today. 

Advances in graphical and 
structural models have made counterfactuals
computationally manageable and thus 
rendered model-driven reasoning a more promising direction
on which to base strong AI.
In the next section, I will describe the impediments facing
machine learning systems using a three-level hierarchy that governs inferences
in causal reasoning.
The final section summarizes how these impediments were
circumvented using modern tools of causal inference.
\newpage

\subsection*{The Three Layer Causal Hierarchy} 
\begin{figure*}[h]
\centering
\begin{tabular}{| p{3.25cm} | p{2.45cm} | p{4.5cm} | p{4.5cm} |}
\hline
Level & Typical & Typical Questions & Examples \\
(Symbol) & Activity & & \\ 
\hline
{1. Association \hfill} \linebreak  $P(y|x)$ & 
Seeing & 
{What is? \hfill}
   \linebreak {How would seeing $X$ \hfill} 
   \linebreak {change my belief in$Y$? \hfill} & 
{What does a symptom tell me about a disease? \hfill}
   \linebreak What does a survey tell us about the election results? \\ \hline

{2. Intervention \hfill} \linebreak $P(y|do(x),z)$ & 
{Doing \linebreak Intervening} & 
{What if?\hfill} 
   \linebreak What if I do $X$? & 
{What if I take aspirin, will my headache be cured? \hfill}
   \linebreak What if we ban cigarettes? \\ \hline 

{3. Counterfactuals \hfill} \linebreak $P(y_{x}|x',y')$ & 
{Imagining, \linebreak Retrospection} & 
{Why? \hfill} 
   \linebreak {Was it $X$ that caused $Y$? \hfill} 
   \linebreak {What if I had acted \hfill} 
   \linebreak differently? &
{Was it the aspirin that stopped my headache? \hfill}
   \linebreak {Would Kennedy be alive had Oswald not shot him? \hfill}
    \linebreak What if I had not been smoking the past 2 years? \\ \hline
\end{tabular}
\caption{The Causal Hierarchy. Questions at level $i$
can only be answered if information from level
$i$ or higher is available. \label{fig1}}
\end{figure*}

An extremely useful insight unveiled by the
logic of causal reasoning is the existence of a sharp
classification of causal information, in terms 
of the kind of questions that each class is 
capable of answering. The classification forms 
a 3-level hierarchy in the sense that questions 
at level $i$ $(i=1,2,3)$ can only be answered if 
information from level $j$ ($j \geq i$) 
is available. 

Figure \ref{fig1} shows the 3-level hierarchy, together with the
characteristic 
questions that can be answered at each level. 
The levels are titled 1.\ Association, 2.\ Intervention, and 3.\
Counterfactual. The names of these layers were chosen to emphasize 
their usage. 
We call the first level Association, because it invokes
purely statistical relationships, defined by the naked data.\footnote{Other names used
for inferences at this layer are: ``model-free,'' 
``model-blind,'' ``black-box,'' or ``data-centric.'' 
\citet{darwiche:17} 
used ``function-fitting,''
for it amounts to fitting data by a complex function
defined by the neural network architecture.}
For instance, observing a customer 
who buys toothpaste makes it more likely that he/she buys floss; such association can be inferred directly
from the observed data using conditional expectation.
Questions at this layer, because they require no causal information, are 
placed at the bottom level on the hierarchy. 
The second level, Intervention, ranks higher than Association 
because it involves not just seeing what is, but changing what we
see. A typical question at this level would be: 
What happens if we double the price? Such questions cannot
be answered from sales data alone, because they involve
a change in customers behavior, in reaction to the new pricing.
These choices may differ substantially from those taken
in previous price-raising situations. (Unless
we replicate precisely the market conditions that existed
when the price reached double its current value.)
Finally, the top level is called Counterfactuals, a term 
that goes back to the philosophers David Hume and John 
Stewart Mill, and which has been given computer-friendly
semantics in the past two decades.  A typical question in the 
counterfactual category is ``What if I had acted differently,'' 
thus necessitating retrospective reasoning. 

Counterfactuals are placed at the top of the hierarchy because 
they subsume interventional and associational questions. 
If we have a model that can answer counterfactual queries, 
we can also answer questions about interventions 
and observations. For example, the interventional question, 
What will happen if we double the price? 
can be answered by asking the counterfactual question: 
What would happen had the price been twice its current value? 
Likewise, associational questions can be answered once we can 
answer interventional questions; we simply ignore the
action part and let observations take over. 
The translation does not work in the opposite direction. 
Interventional questions cannot be answered from purely 
observational information (i.e., from statistical data alone). 
No counterfactual question involving retrospection can be answered 
from purely interventional information, such as that acquired 
from controlled experiments; we cannot re-run an experiment on 
subjects who were treated with a drug and see how they behave
had they not given the drug.
The hierarchy is therefore directional, with the top level 
being the most powerful one. 

Counterfactuals are the building blocks of scientific thinking 
as well as legal and moral reasoning.
In civil court, for example, the defendant is considered to 
be the culprit of an injury if, {\em but for} the 
defendant's action, it is more likely than not that the 
injury would not have occurred. 
The computational meaning of {\em but for} calls for 
comparing the real world to an alternative world in which 
the defendant action did not take place. 

Each layer in the hierarchy has
a syntactic signature that characterizes the
the sentences admitted into that layer. For example, the association 
layer is characterized by conditional probability 
sentences, e.g., $P(y|x)= p$ stating that:
the probability of event $Y=y$ given that we observed
event $X=x$ is equal to $p$. In large systems, such evidential sentences 
can be computed efficiently using Bayesian Networks, 
or any of the neural networks that support 
deep-learning systems. 

At the interventional layer we find sentences 
of the type $P(y|do(x),$\newline$z)$, which denotes
``The probability of event $Y=y$ given that we intervene
and set the value of $X$ to $x$ and subsequently observe event $Z=z$. 
Such expressions can be estimated experimentally from randomized 
trials or analytically using Causal Bayesian 
Networks \citep[Chapter 3]{pearl:2k}. 
A child learns the effects of interventions through
playful manipulation of the environment (usually in a deterministic
playground), and AI planners obtain interventional
knowledge by exercising their designated sets of actions.
Interventional expressions cannot be inferred from 
passive observations alone, regardless of how big the data.

Finally, at the counterfactual level, we have expressions
of the type 
$P(y_{x} |x',y')$ which stand for
``The probability that event $Y=y$ would be observed
had $X$ been $x$, given that 
we actually observed $X$ to be $x'$ and and $Y$ to be $y'$. 
For example, the probability that Joe's salary would
be $y$ had he finished college, given that his actual
salary is $y'$ and that he had only two years of college.''
Such sentences can be computed only when 
we possess functional or Structural Equation models, 
or properties of such models \citep[Chapter 7]{pearl:2k}. 

This hierarchy, and the formal restrictions it entails,
explains why statistics-based 
machine learning systems are prevented from 
reasoning about actions, experiments and 
explanations. It also informs us what extra-statistical information
is needed, and in what format, in order to support those 
modes of reasoning.

Researchers are often surprised that
the hierarchy denegrades the impressive achievements
of deep learning to the level of Association, side
by side with textbook curve-fitting exercises. A popular 
stance against this comparison argues that, whereas the objective
of curve-fitting is to maximize ``fit,''
in deep learning we try to minimize ``over fit.''
Unfortunately, the theoretical barriers that separate
the three layers in the hierarchy tell us that the
nature of our objective function does not matter.
As long as our system optimizes some property
of the observed data, however noble or sophisticated,
while making no reference to the world
outside the data, we are back to level-1 of the hierarchy 
with all the limitations that this level entails.



\subsection*{The Seven Pillars of the Causal Revolution 
(or What you can do with a causal model that you could not
do without?)}

Consider the following five questions:
\begin{itemize}
\item How effective is a given treatment in preventing a disease?
\item Was it the new tax break that caused our sales to go up?
\item What is the annual health-care costs attributed to obesity?
\item Can hiring records prove an employer guilty of sex discrimination?
\item I am about to quit my gob, but should I?
\end{itemize}

The common feature of these questions is that they 
are concerned with cause-and-effect relationships. We can
recognize them through words such as ``preventing,'' ``cause,''
``attributed to,'' ``discrimination,'' and ``should I.''
Such words are common in everyday language, and our society
constantly demands answers to such questions.
Yet, until very recently science gave us no means even to
articulate them, let alone answer them.
Unlike the rules of geometry, mechanics, optics or probabilities, 
the rules of cause and effect have been denied the benefits of mathematical analysis.

To appreciate the extent of this denial, 
readers would be stunned to know that
only a few decades ago scientists were unable to
write down a mathematical equation for the obvious fact that
``mud does not cause rain.'' Even today, only the top
echelon of the scientific community can write such an
equation and formally distinguish ``mud causes rain'' from 
``rain causes mud.'' And you would probably be even more
surprised to discover that your favorite college professor is 
not among them. 

Things have changed dramatically in the past three decades,
A mathematical
language has been developed for managing causes and effects,
accompanied by a set of tools that turn causal analysis
into a mathematical game, not unlike solving algebraic
equations, or finding proofs in high-school geometry.
These tools permit us to express causal questions formally 
codify our existing knowledge in both diagrammatic
and algebraic forms, and then leverage our data to 
estimate the answers. Moreover, the theory warns us when
the state of existing knowledge or the available data
are insufficient to answer our questions; and then 
suggests additional sources of knowledge or data to make the 
questions answerable.

Harvard professor Garry King gave this
transformation a historical perspective:
``More has been learned about causal inference in the last
few decades than the sum total of everything that had been
learned about it in all prior recorded history'' \citep{morgan:win15}. 
I call this transformation ``The Causal Revolution,'' \citep{pearl:mac18}
and the mathematical framework that led to it
I call ``Structural Causal Models (SCM).''

The SCM deploys three parts
\begin{enumerate}
\item Graphical models, 
\item Structural equations, and
\item Counterfactual and interventional logic
\end{enumerate}
Graphical models serve as a language
for representing what we know about the world,
counterfactuals help us to articulate what we
want to know, while structural equations serve to 
tie the two together in a solid semantics.

\begin{figure}[h] 
\begin{center} 
\epsfig{figure=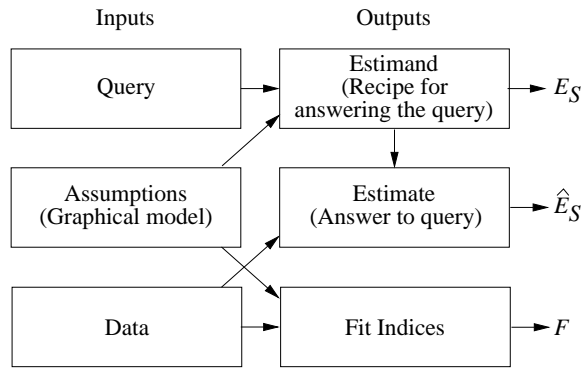,width=3in}
\end{center}
\caption{How the SCM ``inference engine'' combines data with causal model (or assumptions) to produce 
answers to queries of interest.\label{fig2}}
\end{figure} 
Figure \ref{fig2} illustrates the operation of SCM in the form of
an inference engine. The engine accepts three inputs:
Assumptions, Queries, and Data, and produces three outputs:
Estimand, Estimate and Fit indices. The Estimand $(E_S)$ is a 
mathematical formula that, based on the Assumptions, provides
a recipe for answering the Query from any hypothetical data, 
whenever they are available. After receiving the Data, the
engine uses the Estimand to produce an actual Estimate
$(\hat{E}_S)$
for the answer, along with statistical estimates of the 
confidence in that answer (To reflect the limited size of
the data set, as well as possible measurement errors or
missing data.) Finally, the engine produces a list of
``fit indices'' which measure how compatible the data are
with the Assumptions conveyed by the model.
 
To exemplify these operations, let us assume that our Query 
stands for the causal effect of $X$ on $Y$, written $Q=P(Y|do(X))$, 
where $X$ and $Y$ are two variables of interest. 
Let the modeling assumptions be encoded in the graph below,
\begin{figure}[h] 
\begin{center} 
\epsfig{figure=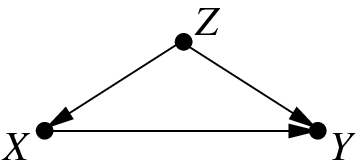,width=1.25in}
\end{center}
\end{figure}
where $Z$ is a third variable affecting both $X$ and $Y$. 
Finally, let the data
be sampled at random from a joint distribution $P(X,Y,Z)$.
The Estimand $(E_S)$ calculated by the engine will be the
formula $E_S = \sum_z P(Y|X,Z)P(Z)$. It defines
a property of P(X,Y,Z) that, if estimated,
would provide a correct answer to our Query. The answer itself,
the Estimate $\hat{E}_S$, can be produced by any number of techniques
that produce a consistent estimate of $E_S$
from finite samples of $P(X,Y,Z)$. For example, the sample
average (of $Y$) over all cases satisfying the specified
$X$ and $Z$ conditions, would be a consistent estimate.
But more efficient estimation techniques can be devised to
overcome data sparsity \citep{rosenbaum:rub83}. 
This is where deep learning excels and where most work in machine learning
has been focused, albeit with no guidance of a model-based estimand. Finally, the Fit
Index in our example will be NULL. In other words,
after examining the structure of the graph, the
engine should conclude that the assumptions encoded do not
have any testable implications. Therefore, the veracity of
resultant estimate must lean entirely on the
assumptions encoded in the graph -- no refutation
nor corroboration can be obtained from the data.\footnote{The assumptions encoded in the graph
are conveyed by its missing arrows. For example,
$Y$ does not influence $X$ or $Z$, $X$ does not influence $Z$
and, most importantly, $Z$ is the only variable affecting
both $X$ and $Y$. That these assumptions lack testable implications can
be concluded from the fact that the graph is complete, i.e.,
no edges are missing.}
 
The same procedure applies to more sophisticated
queries, for example, the counterfactual query $Q=P(y_x|x',y')$
discussed before. We may also permit some of the data to 
arrive from controlled experiments, which would take the form
$P(V|do(W))$, in case W is the controlled variable.
The role of the Estimand would remain that
of converting the Query into the syntactic format of the 
available data and, then, guiding the choice of the estimation
technique to ensure unbiased estimates. Needless to state, 
the conversion task is not always feasible, in which case the Query will
be declared ``non-identifiable'' and the engine should
exit with FAILURE. Fortunately, efficient and complete 
algorithms have been developed to decide identifiability 
and to produce estimands for a variety of counterfactual
queries and a variety of data types \citep{bareinboim:pea16-r450}. 

Next we provide a bird's eye view of seven
accomplishments of the SCM framework and discuss the unique 
contribution that each pillar brings to the art of automated reasoning.


\section*{Pillar 1: Encoding Causal Assumptions -- Transparency and Testability \label{sec1}}

The task of encoding assumptions in a compact and
usable form, is not a trivial matter once we take 
seriously the requirement of transparency and testability.\footnote{Economists, for example, having chosen algebraic
over graphical representations, are deprived of elementary testability-detecting features \citep{pearl:15-r391}.} 
Transparency enables analysts to discern whether the assumptions encoded
are plausible (on scientific grounds), or whether additional 
assumptions are warranted.
Testability permits us (be it an analyst or a machine) to
determine whether the assumptions encoded are compatible
with the available data and, if not, identify those
that need repair. 

Advances in graphical models have
made compact encoding feasible. Their transparency 
stems naturally from the fact that all assumptions are encoded
graphically, mirroring the way researchers perceive of
cause-effect relationship in the domain;
judgments of counterfactual or
statistical dependencies are not required, since these can 
be read off the structure of the graph.
Testability is facilitated through
a graphical criterion called $d$-separation, which
provides the fundamental connection between
causes and probabilities.
It tells us, for any given pattern
of paths in the model, what pattern of dependencies we
should expect to find in the data \citep{pearl:88a}. 

\section*{Pillar 2: $Do$-calculus and the control of confounding \label{sec2}}

Confounding, or the presence of unobserved causes of two or
more variables, has long been consider the the major obstacle to
drawing causal inference from data, This obstacle had been 
demystified and ``deconfounded'' through a graphical
criterion called ``back-door.'' In particular, the task of 
selecting an appropriate set of covariates to control for
confounding has been reduced to a simple ``roadblocks''
puzzle manageable by a simple algorithm \citep{pearl:93ss}. 

For models where the ``back-door'' criterion does not hold,
a symbolic engine is available, called $do$-{\em calculus}, 
which predicts the effect of policy interventions 
whenever feasible, and exits with failure whenever
predictions cannot be ascertained with the specified
assumptions \citep{pearl:95,tian:pea02,shpitser:pea08-r336}.  


\section*{Pillar 3: The Algorithmization of Counterfactuals \label{sec3}}

Counterfactual analysis deals with behavior of specific
individuals, identified by a distinct set of characteristics,
For example, given that Joe's salary is $Y = y$, and that he 
went $X=x$ years to college, what would Joe's salary be had he 
had one more year of education.

One of the crown achievements of the Causal Revolution has been 
to formalize counterfactual reasoning
within the graphical representation, the
very representation researchers use to encode scientific knowledge.
Every structural equation model determines
the truth value of every counterfactual
sentence. Therefore, we can determine analytically 
if the probability of the sentence is estimable from experimental
or observational studies, or combination thereof [\citealp{balke:pea94a}; \citealp[Chapter 7]{pearl:2k}]. 

Of special interest in causal discourse are counterfactual
questions concerning ``causes of effects,'' as opposed to
``effects of causes.''  For example, how likely it is that 
Joe's swimming exercise was a necessary (or 
sufficient) cause of Joe's death \citep{pearl:15-r431,halpern:pea05-r266a}. 



\section*{Pillar 4: Mediation Analysis and the Assessment of Direct and
Indirect Effects \label{sec4}}

Mediation analysis concerns the mechanisms 
that transmit changes from a cause to its effects.
The identification of such intermediate
mechanism is essential for generating explanations and 
counterfactual analysis must be invoked to facilitate
this identification. The graphical 
representation of counterfactuals enables us to define 
direct and indirect effects and to decide when
these effects are estimable from data, or experiments \citep{robins:gre92, pearl:01,vanderweele:15}.
Typical queries answerable by this analysis are: What
fraction of the effect of $X$ on $Y$ is mediated by variable $Z$.

\section*{Pillar 5: External Validity and Sample Selection Bias \label{sec5}}
The validity of every experimental study is challenged
by disparities between the experimental and implementational
setups.  A machine trained in one environment cannot
be expected to perform well when environmental conditions
change, unless the changes are localized and identified.
This problem, and its various manifestations are well
recognized by machine-learning researchers, and enterprises
such as ``domain adaptation,'' ``transfer learning,'' 
``life-long learning,'' and ``explainable AI,'' are just some 
of the subtasks identified by researchers and funding agencies in an
attempt to alleviate the general problem of robustness.
Unfortunately, the problem of robustness requires a
causal model of the environment, and cannot be handled at the
level of Association, in which most remedies were tried.
Associations are not sufficient for identifying
the mechanisms affected by changes that occurred.
The $do$-calculus discussed above now offers
a complete methodology for overcoming bias due
to environmental changes.  It can be used both 
for re-adjusting learned policies to circumvent environmental 
changes and for controlling bias due to non-representative 
samples \citep{bareinboim:pea16-r450}. 

\section*{Pillar 6: Missing Data \label{sec6}} 

Problems of missing data plague every branch of experimental
science. Respondents do not answer every item on a
questionnaire, sensors fade as environmental conditions change,
and patients often drop from a clinical study for
unknown reasons. The rich literature on this problem is
wedded to a model-blind paradigm of statistical analysis
and, accordingly, it is severely limited to situations where
missingness occurs at random, that is, independent of values
taken by other variables in the model. Using causal models of the 
missingness process we can now formalize the conditions under which
causal and probabilistic relationships can be recovered from 
incomplete data and, whenever the conditions are satisfied,
produce a consistent estimate of the desired relationship \citep{mohan:pea17-r473}. 

\section*{Pillar 7: Causal Discovery \label{sec7}}

The $d$-separation criterion described above
enables us to detect and enumerate the testable
implications of a given causal model. This opens
the possibility of inferring, with mild assumptions,
the set of models that are compatible with
the data, and to represent this set compactly.
Systematic searches have been developed which, 
in certain circumstances,
can prune the set of compatible models significantly
to the point where causal queries can be
estimated directly from that set \citep{spirtes:etal00,pearl:2k,peters:etal17}. 

\section*{Conclusions}
The philosopher Stephen Toulmin (\citeyear{toulmin:61}) 
identifies model-based vs.\ model-blind dichotomy as the key
to understanding the ancient rivalry between Babylonian
and Greek science.  According to Toulmin,
the Babylonians astronomers were masters of black-box
prediction, far surpassing their Greek rivals in accuracy
and consistency \cite[pp.\ 27--30]{toulmin:61}. 
Yet Science favored the 
creative-speculative  strategy of the Greek astronomers which was 
wild with metaphysical imagery: circular tubes full of fire, 
small holes through which celestial fire was visible as stars,
and hemispherical earth riding on turtle backs.
Yet it was this wild modeling strategy, not Babylonian rigidity,
that jolted Eratosthenes (276-194 BC) to perform one of the most
creative  experiments in the ancient world and measure the 
radius of the earth. This would never have occurred to
a Babylonian curve-fitter.

Coming back to strong AI, we have seen that
model-blind approaches have intrinsic
limitations on the cognitive tasks that they can perform.
We have described some of these tasks and demonstrated
how they can be accomplished in the SCM framework, and why a model-based approach
is essential for performing these tasks. Our general conclusion is that human-level AI cannot emerge 
solely from model-blind learning machines; 
it requires the symbiotic collaboration of data and models.

Data science is only as much of a science 
as it facilitates the interpretation of data --
a two-body problem, connecting data to reality.
Data alone are hardly a science, regardless how big they get and 
how skillfully they are manipulated.

\section*{Acknowledgement}

This research was supported in parts by grants from Defense Advanced Research Projects Agency [\#W911NF-16-057], National Science Foundation [\#IIS-1302448, \#IIS-1527490, and \#IIS-1704932], and Office of Naval Research [\#N00014-17-S-B001].



\begin{thebibliography}{22}


\ifx \showCODEN    \undefined \def \showCODEN     #1{\unskip}     \fi
\ifx \showDOI      \undefined \def \showDOI       #1{#1}\fi
\ifx \showISBNx    \undefined \def \showISBNx     #1{\unskip}     \fi
\ifx \showISBNxiii \undefined \def \showISBNxiii  #1{\unskip}     \fi
\ifx \showISSN     \undefined \def \showISSN      #1{\unskip}     \fi
\ifx \showLCCN     \undefined \def \showLCCN      #1{\unskip}     \fi
\ifx \shownote     \undefined \def \shownote      #1{#1}          \fi
\ifx \showarticletitle \undefined \def \showarticletitle #1{#1}   \fi
\ifx \showURL      \undefined \def \showURL       {\relax}        \fi
\providecommand\bibfield[2]{#2}
\providecommand\bibinfo[2]{#2}
\providecommand\natexlab[1]{#1}
\providecommand\showeprint[2][]{arXiv:#2}

\bibitem[\protect\citeauthoryear{Balke and Pearl}{Balke and Pearl}{1994}]%
        {balke:pea94a}
\bibfield{author}{\bibinfo{person}{A. Balke} {and} \bibinfo{person}{J. Pearl}.}
  \bibinfo{year}{1994}\natexlab{}.
\newblock \showarticletitle{Probabilistic evaluation of counterfactual
  queries}.
\newblock In \bibinfo{booktitle}{\emph{Proceedings of the Twelfth National
  Conference on Artificial Intelligence}}. Vol.~\bibinfo{volume}{I}.
  \bibinfo{publisher}{MIT Press}, \bibinfo{address}{Menlo Park, CA},
  \bibinfo{pages}{230--237}.
\newblock


\bibitem[\protect\citeauthoryear{Bareinboim and Pearl}{Bareinboim and
  Pearl}{2016}]%
        {bareinboim:pea16-r450}
\bibfield{author}{\bibinfo{person}{E.\ Bareinboim} {and} \bibinfo{person}{J.
  Pearl}.} \bibinfo{year}{2016}\natexlab{}.
\newblock \showarticletitle{Causal inference and the data-fusion problem}.
\newblock \bibinfo{journal}{\emph{Proceedings of the National Academy of
  Sciences}}  \bibinfo{volume}{113} (\bibinfo{year}{2016}),
  \bibinfo{pages}{7345--7352}.
\newblock
Issue 27.


\bibitem[\protect\citeauthoryear{Darwiche}{Darwiche}{2017}]%
        {darwiche:17}
\bibfield{author}{\bibinfo{person}{A. Darwiche}.}
  \bibinfo{year}{2017}\natexlab{}.
\newblock \bibinfo{booktitle}{\emph{Human-Level Intelligence or Animal-Like
  Abilities?}}
\newblock \bibinfo{type}{{T}echnical {R}eport}.
  \bibinfo{institution}{Department of Computer Science, University of
  California, Los Angeles}, \bibinfo{address}{CA}.
\newblock
\newblock
\shownote{arXiv:1707.04327.}


\bibitem[\protect\citeauthoryear{Halpern and Pearl}{Halpern and Pearl}{2005}]%
        {halpern:pea05-r266a}
\bibfield{author}{\bibinfo{person}{J.Y. Halpern} {and} \bibinfo{person}{J.
  Pearl}.} \bibinfo{year}{2005}\natexlab{}.
\newblock \showarticletitle{Causes and Explanations: A Structural-Model
  Approach---{Part I}: Causes}.
\newblock \bibinfo{journal}{\emph{British Journal of Philosophy of Science}}
  \bibinfo{volume}{56} (\bibinfo{year}{2005}), \bibinfo{pages}{843--887}.
\newblock


\bibitem[\protect\citeauthoryear{Mohan and Pearl}{Mohan and Pearl}{2017}]%
        {mohan:pea17-r473}
\bibfield{author}{\bibinfo{person}{K. Mohan} {and} \bibinfo{person}{J. Pearl}.}
  \bibinfo{year}{2017}\natexlab{}.
\newblock \bibinfo{booktitle}{\emph{Graphical Models for Processing Missing
  Data}}.
\newblock \bibinfo{type}{{T}echnical {R}eport} R-473,
  {$<$http://ftp.cs.ucla.edu/pub/stat\_ser/r473.pdf$>$}.
  \bibinfo{institution}{Department of Computer Science, University of
  California, Los Angeles}, \bibinfo{address}{CA}.
\newblock
\newblock
\shownote{Submitted.}


\bibitem[\protect\citeauthoryear{Morgan and Winship}{Morgan and
  Winship}{2015}]%
        {morgan:win15}
\bibfield{author}{\bibinfo{person}{S.L. Morgan} {and} \bibinfo{person}{C.
  Winship}.} \bibinfo{year}{2015}\natexlab{}.
\newblock \bibinfo{booktitle}{\emph{Counterfactuals and Causal Inference:
  Methods and Principles for Social Research (Analytical Methods for Social
  Research)} (\bibinfo{edition}{2nd} ed.)}.
\newblock \bibinfo{publisher}{Cambridge University Press},
  \bibinfo{address}{New York, NY}.
\newblock


\bibitem[\protect\citeauthoryear{Pearl}{Pearl}{1988}]%
        {pearl:88a}
\bibfield{author}{\bibinfo{person}{J. Pearl}.} \bibinfo{year}{1988}\natexlab{}.
\newblock \bibinfo{booktitle}{\emph{Probabilistic Reasoning in Intelligent
  Systems}}.
\newblock \bibinfo{publisher}{Morgan Kaufmann}, \bibinfo{address}{San Mateo,
  CA}.
\newblock


\bibitem[\protect\citeauthoryear{Pearl}{Pearl}{1993}]%
        {pearl:93ss}
\bibfield{author}{\bibinfo{person}{J. Pearl}.} \bibinfo{year}{1993}\natexlab{}.
\newblock \showarticletitle{Comment: Graphical Models, Causality, and
  Intervention}.
\newblock \bibinfo{journal}{\emph{Statist. Sci.}} \bibinfo{volume}{8},
  \bibinfo{number}{3} (\bibinfo{year}{1993}), \bibinfo{pages}{266--269}.
\newblock


\bibitem[\protect\citeauthoryear{Pearl}{Pearl}{1995}]%
        {pearl:95}
\bibfield{author}{\bibinfo{person}{J. Pearl}.} \bibinfo{year}{1995}\natexlab{}.
\newblock \showarticletitle{Causal diagrams for empirical research}.
\newblock \bibinfo{journal}{\emph{Biometrika}} \bibinfo{volume}{82},
  \bibinfo{number}{4} (\bibinfo{year}{1995}), \bibinfo{pages}{669--710}.
\newblock


\bibitem[\protect\citeauthoryear{Pearl}{Pearl}{2000}]%
        {pearl:2k}
\bibfield{author}{\bibinfo{person}{J. Pearl}.} \bibinfo{year}{2000}\natexlab{}.
\newblock \bibinfo{booktitle}{\emph{Causality: Models, Reasoning, and
  Inference}}.
\newblock \bibinfo{publisher}{Cambridge University Press},
  \bibinfo{address}{New York}.
\newblock
\newblock
\shownote{2nd edition, 2009.}


\bibitem[\protect\citeauthoryear{Pearl}{Pearl}{2001}]%
        {pearl:01}
\bibfield{author}{\bibinfo{person}{J. Pearl}.} \bibinfo{year}{2001}\natexlab{}.
\newblock \showarticletitle{Direct and indirect effects}.
\newblock In \bibinfo{booktitle}{\emph{Uncertainty in Artificial Intelligence,
  Proceedings of the Seventeenth Conference}}. \bibinfo{publisher}{Morgan
  Kaufmann}, \bibinfo{address}{San Francisco, CA}, \bibinfo{pages}{411--420}.
\newblock


\bibitem[\protect\citeauthoryear{Pearl}{Pearl}{2015a}]%
        {pearl:15-r431}
\bibfield{author}{\bibinfo{person}{J. Pearl}.}
  \bibinfo{year}{2015}\natexlab{a}.
\newblock \showarticletitle{Causes of Effects and Effects of Causes}.
\newblock \bibinfo{journal}{\emph{Journal of Sociological Methods and
  Research}}  \bibinfo{volume}{44} (\bibinfo{year}{2015}),
  \bibinfo{pages}{149--164}.
\newblock
Issue 1.


\bibitem[\protect\citeauthoryear{Pearl}{Pearl}{2015b}]%
        {pearl:15-r391}
\bibfield{author}{\bibinfo{person}{J. Pearl}.}
  \bibinfo{year}{2015}\natexlab{b}.
\newblock \showarticletitle{{Trygve Haavelmo} and the emergence of causal
  calculus}.
\newblock \bibinfo{journal}{\emph{Econometric Theory}}  \bibinfo{volume}{31}
  (\bibinfo{year}{2015}), \bibinfo{pages}{152--179}.
\newblock
Issue 1.
\newblock
\shownote{Special issue on {Haavelmo} Centennial.}


\bibitem[\protect\citeauthoryear{Pearl and Mackenzie}{Pearl and
  Mackenzie}{2018}]%
        {pearl:mac18}
\bibfield{author}{\bibinfo{person}{J.\ Pearl} {and} \bibinfo{person}{D.
  Mackenzie}.} \bibinfo{year}{2018, forthcoming}\natexlab{}.
\newblock \bibinfo{booktitle}{\emph{The Book of Why: The New Science of Cause
  and Effect}}.
\newblock \bibinfo{publisher}{Basic Books}, \bibinfo{address}{New York}.
\newblock


\bibitem[\protect\citeauthoryear{Peters, Janzing, and {Sch\"{o}lkopf}}{Peters
  et~al\mbox{.}}{2017}]%
        {peters:etal17}
\bibfield{author}{\bibinfo{person}{J.\ Peters}, \bibinfo{person}{D.\ Janzing},
  {and} \bibinfo{person}{B. {Sch\"{o}lkopf}}.} \bibinfo{year}{2017}\natexlab{}.
\newblock \bibinfo{booktitle}{\emph{Elements of Causal Inference -- Foundations
  and Learning Algorithms}}.
\newblock \bibinfo{publisher}{The MIT Press}, \bibinfo{address}{Cambridge, MA}.
\newblock


\bibitem[\protect\citeauthoryear{Robins and Greenland}{Robins and
  Greenland}{1992}]%
        {robins:gre92}
\bibfield{author}{\bibinfo{person}{J.M. Robins} {and} \bibinfo{person}{S.
  Greenland}.} \bibinfo{year}{1992}\natexlab{}.
\newblock \showarticletitle{Identifiability and Exchangeability for Direct and
  Indirect Effects}.
\newblock \bibinfo{journal}{\emph{Epidemiology}} \bibinfo{volume}{3},
  \bibinfo{number}{2} (\bibinfo{year}{1992}), \bibinfo{pages}{143--155}.
\newblock


\bibitem[\protect\citeauthoryear{Rosenbaum and Rubin}{Rosenbaum and
  Rubin}{1983}]%
        {rosenbaum:rub83}
\bibfield{author}{\bibinfo{person}{P. Rosenbaum} {and} \bibinfo{person}{D.
  Rubin}.} \bibinfo{year}{1983}\natexlab{}.
\newblock \showarticletitle{The central role of propensity score in
  observational studies for causal effects}.
\newblock \bibinfo{journal}{\emph{Biometrika}}  \bibinfo{volume}{70}
  (\bibinfo{year}{1983}), \bibinfo{pages}{41--55}.
\newblock


\bibitem[\protect\citeauthoryear{Shpitser and Pearl}{Shpitser and
  Pearl}{2008}]%
        {shpitser:pea08-r336}
\bibfield{author}{\bibinfo{person}{I.\ Shpitser} {and} \bibinfo{person}{J.
  Pearl}.} \bibinfo{year}{2008}\natexlab{}.
\newblock \showarticletitle{Complete Identification Methods for the Causal
  Hierarchy}.
\newblock \bibinfo{journal}{\emph{Journal of Machine Learning Research}}
  \bibinfo{volume}{9} (\bibinfo{year}{2008}), \bibinfo{pages}{1941--1979}.
\newblock


\bibitem[\protect\citeauthoryear{Spirtes, Glymour, and Scheines}{Spirtes
  et~al\mbox{.}}{2000}]%
        {spirtes:etal00}
\bibfield{author}{\bibinfo{person}{P. Spirtes}, \bibinfo{person}{C.N. Glymour},
  {and} \bibinfo{person}{R. Scheines}.} \bibinfo{year}{2000}\natexlab{}.
\newblock \bibinfo{booktitle}{\emph{Causation, Prediction, and Search}
  (\bibinfo{edition}{2nd} ed.)}.
\newblock \bibinfo{publisher}{MIT Press}, \bibinfo{address}{Cambridge, MA}.
\newblock


\bibitem[\protect\citeauthoryear{Tian and Pearl}{Tian and Pearl}{2002}]%
        {tian:pea02}
\bibfield{author}{\bibinfo{person}{J.\ Tian} {and} \bibinfo{person}{J. Pearl}.}
  \bibinfo{year}{2002}\natexlab{}.
\newblock \showarticletitle{A general identification condition for causal
  effects}.
\newblock In \bibinfo{booktitle}{\emph{Proceedings of the Eighteenth National
  Conference on Artificial Intelligence}}. \bibinfo{publisher}{AAAI Press/The
  MIT Press}, \bibinfo{address}{Menlo Park, CA}, \bibinfo{pages}{567--573}.
\newblock


\bibitem[\protect\citeauthoryear{Toulmin}{Toulmin}{1961}]%
        {toulmin:61}
\bibfield{author}{\bibinfo{person}{S. Toulmin}.}
  \bibinfo{year}{1961}\natexlab{}.
\newblock \bibinfo{booktitle}{\emph{Forecast and Understanding}}.
\newblock \bibinfo{publisher}{University Press}, \bibinfo{address}{Indiana}.
\newblock


\bibitem[\protect\citeauthoryear{{VanderWeele}}{{VanderWeele}}{2015}]%
        {vanderweele:15}
\bibfield{author}{\bibinfo{person}{T.J. {VanderWeele}}.}
  \bibinfo{year}{2015}\natexlab{}.
\newblock \bibinfo{booktitle}{\emph{Explanation in Causal Inference: Methods
  for Mediation and Interaction}}.
\newblock \bibinfo{publisher}{Oxford University Press}, \bibinfo{address}{New
  York}.
\newblock


\end{thebibliography}


\end{document}